\RequirePackage[l2tabu, orthodox]{nag}
\documentclass[11pt]{article}

\usepackage[T1]{fontenc}

\usepackage[bitstream-charter]{mathdesign}
\usepackage{amsmath}
\usepackage[scaled=0.92]{PTSans}

\usepackage[
  paper  = letterpaper,
  left   = 1.65in,
  right  = 1.65in,
  top    = 1.0in,
  bottom = 1.0in,
  ]{geometry}

\usepackage[usenames,dvipsnames]{xcolor}
\definecolor{shadecolor}{gray}{0.9}

\usepackage[final,expansion=alltext]{microtype}
\usepackage[english]{babel}
\usepackage[parfill]{parskip}
\usepackage{afterpage}
\usepackage{framed}

\usepackage{lineno}

\usepackage{ragged2e}

\newcounter{parcount}

\usepackage{booktabs}
\usepackage{multirow}
\usepackage{tabularx}

\usepackage{listings}
\usepackage{fancyvrb}
\fvset{fontsize=\normalsize}

\usepackage{natbib}

\usepackage[colorlinks,linktoc=all]{hyperref}
\usepackage[all]{hypcap}
\hypersetup{citecolor=BurntOrange}
\hypersetup{linkcolor=black}
\hypersetup{urlcolor=MidnightBlue}

\usepackage[acronym,nowarn]{glossaries}

\lstdefinestyle{mystyle}{
    commentstyle=\color{OliveGreen},
    keywordstyle=\color{BurntOrange},
    numberstyle=\tiny\color{black!60},
    stringstyle=\color{MidnightBlue},
    basicstyle=\ttfamily,
    breakatwhitespace=false,
    breaklines=true,
    captionpos=b,
    keepspaces=true,
    numbers=left,
    numbersep=5pt,
    showspaces=false,
    showstringspaces=false,
    showtabs=false,
    tabsize=2
}
\usepackage[colorinlistoftodos,
           prependcaption,
           textsize=small,
           backgroundcolor=yellow,
           linecolor=lightgray,
           bordercolor=lightgray]{todonotes}

\usepackage{amsthm}  

\usepackage{bm}
\usepackage[Export]{adjustbox}

\usepackage[colorinlistoftodos,
           prependcaption,
           textsize=small,
           backgroundcolor=yellow,
           linecolor=lightgray,
           bordercolor=lightgray]{todonotes}

\usepackage{graphicx}
\usepackage{caption}
\usepackage{subcaption}
\usepackage{wrapfig}

\usepackage{booktabs}
\usepackage{arydshln} \usepackage{multirow}

\usepackage{listings}
\usepackage{fancyvrb}
\fvset{fontsize=\normalsize}

\hypersetup{linkcolor=black}
\hypersetup{urlcolor=BurntOrange}

\usepackage{listings}
\lstdefinestyle{alp_style}{
    commentstyle=\color{OliveGreen},
    numberstyle=\tiny\color{black!60},
    stringstyle=\color{BrickRed},
    basicstyle=\ttfamily\scriptsize,
    breakatwhitespace=false,
    breaklines=true,
    captionpos=b,
    keepspaces=true,
    numbers=none,
    numbersep=5pt,
    showspaces=false,
    showstringspaces=false,
    showtabs=false,
    tabsize=2
}

\usepackage{capt-of}

\usepackage{lipsum}

\usepackage{authblk}
\usepackage{enumitem}

\theoremstyle{remark}
\newtheorem*{lemma*}{Lemma}

\newcommand{\resultalt}[2]{\makebox[3em]{\hfill#1}\makebox[3.2em]{\hfill(#2)}}
\newcommand{\resultboldalt}[2]{\makebox[3em]{\hfill\textbf{#1}}\makebox[3.2em]{\hfill\textbf{(#2)}}}

\usepackage{algorithm}
\usepackage{algpseudocode}
\usepackage{rotating}

\DeclareMathOperator*{\argmax}{arg\,max}

\def\eqref#1{equation~\ref{#1}}

\def\1{\bm{1}}

\DeclareMathAlphabet{\mathsfit}{\encodingdefault}{\sfdefault}{m}{sl}
\SetMathAlphabet{\mathsfit}{bold}{\encodingdefault}{\sfdefault}{bx}{n}

 \newacronym{ALI}{ali}{adversarially learned inference}
\newacronym{BIGAN}{bigan}{bidirectional generative adversarial network}
\newacronym{VI}{vi}{variational inference}
\newacronym{KL}{kl}{Kullback-Leibler}
\newacronym{ELBO}{elbo}{evidence lower bound}
\newacronym{MCMC}{mcmc}{Markov chain Monte Carlo}
\newacronym{HMC}{hmc}{Hamiltonian Monte Carlo}
\newacronym{RNN}{rnn}{recurrent neural network}
\newacronym{MLP}{mlp}{feed forward neural network}
\newacronym{GAN}{gan}{generative adversarial network}
\newacronym{DCGAN}{dcgan}{deep convolutional generative adversarial network}
\newacronym{PresGAN}{presgan}{prescribed generative adversarial network}
\newacronym{DGM}{dgm}{deep generative model}
\newacronym{PGAN}{pgan}{prescribed generative adversarial network}
\newacronym{VEEGAN}{veegan}{vee {GAN}}
\newacronym{PACGAN}{pacgan}{packed {GAN}}
\newacronym{STYLEGAN}{stylegan}{Style {GAN}}
\newacronym{FID}{fid}{{F}r\'{e}chet {I}nception distance}
\newacronym{IS}{is}{{I}nception score}
\newacronym{ML}{ml}{machine learning}
\newacronym{VS}{vs}{vendi score}
\newacronym{NLP}{nlp}{natural language processing}
\newacronym{IntDiv}{intdiv}{{I}nternal {D}iversity}
\newacronym{BLEU}{bleu}{BLEU}
\newacronym{PAIRWISE-BLEU}{pairwise-bleu}{PAIRWISE-BLEU}
\newacronym{D-LEX-SIM}{d-lex-sim}{D-LEX-SIM}
\newacronym{GILBO}{gilbo}{GILBO}
\newacronym{NOM}{nom}{number of modes}
\newacronym{HMM}{hmm}{HMM}
\newacronym{AAE}{aae}{AAE}
\newacronym{VAE}{vae}{VAE}
\newacronym{JTN}{jtn}{JTN}
\newacronym{Char-RNN}{char-rnn}{Char-RNN}
\newacronym{SMILES}{smiles}{SMILES}
\newacronym{MNIST}{mnist}{MNIST}
\newacronym{MultiNLI}{multinli}{MultiNLI}
\newacronym{StackedMNIST}{stackedmnist}{StackedMNIST}
\newacronym{NLI}{nli}{NLI}
\newacronym{VDVAE}{vdvae}{VDVAE}
\newacronym{LSUN}{lsun}{LSUN}
\newacronym{CIFAR}{cifar}{CIFAR}

\title{\textbf{Quality-Weighted Vendi Scores and\\Their Application to Diverse Experimental Design}}

\author[1]{Quan Nguyen}
\author[2,3, *]{Adji Bousso Dieng}
\affil[1]{Department of Computer Science \& Engineering \protect\\ Washington University in St.\ Louis}
\affil[2]{Department of Computer Science, Princeton University}
\affil[3]{\href{https://vertaix.princeton.edu/}{Vertaix}}
\affil[*]{Published in \emph{International Conference on Machine Learning}, ICML 2024.}

\begin{document}
\maketitle

\begin{abstract}
\noindent
Experimental design techniques such as active search and Bayesian optimization are widely used in the natural sciences for data collection and discovery. However, existing techniques tend to favor exploitation over exploration of the search space, which causes them to get stuck in local optima. This \emph{collapse} problem prevents experimental design algorithms from yielding diverse high-quality data. In this paper, we extend the Vendi scores---a family of interpretable similarity-based diversity metrics---to account for quality. We then leverage these \emph{quality-weighted Vendi scores} to tackle experimental design problems across various applications, including drug discovery, materials discovery, and reinforcement learning. We found that quality-weighted Vendi scores allow us to construct policies for experimental design that flexibly balance quality and diversity, and ultimately assemble rich and diverse sets of high-performing data points. Our algorithms led to a 70\%--170\% increase in the number of effective discoveries compared to baselines.\footnote{Code can be found at \url{https://github.com/vertaix/Quality-Weighted-Vendi-Score}.}\\

\end{abstract}

\section{Introduction}
\glsresetall

\begin{figure*}[t]
\setlength\tabcolsep{6pt}
\adjustboxset{width=\textwidth,valign=c}
\centering
\begin{tabularx}{1.0\textwidth}{@{}
  l
  X @{\hspace{6pt}}
  X @{\hspace{6pt}}
  X @{\hspace{6pt}}
  X @{\hspace{6pt}}
  X
@{}}
\rotatebox[origin=c]{0}{VS}
& & \includegraphics[scale=0.18]{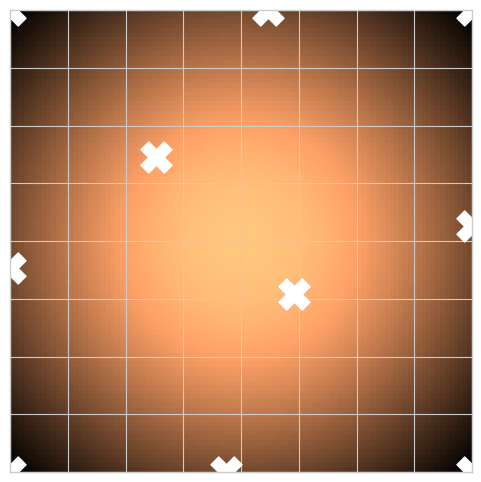}
& \includegraphics[scale=0.18]{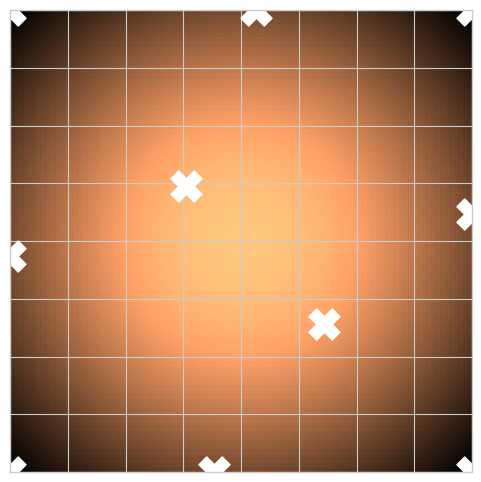}
& \includegraphics[scale=0.18]{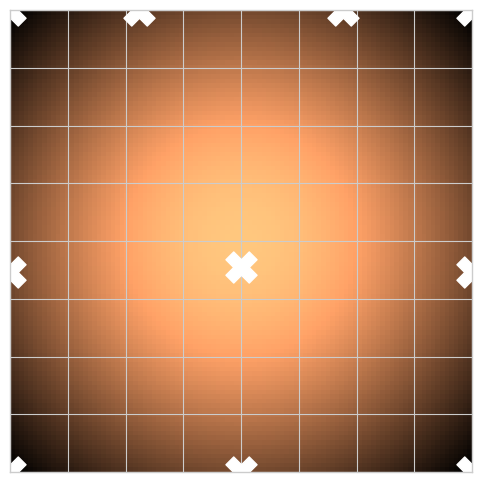}
& \includegraphics[scale=0.18]{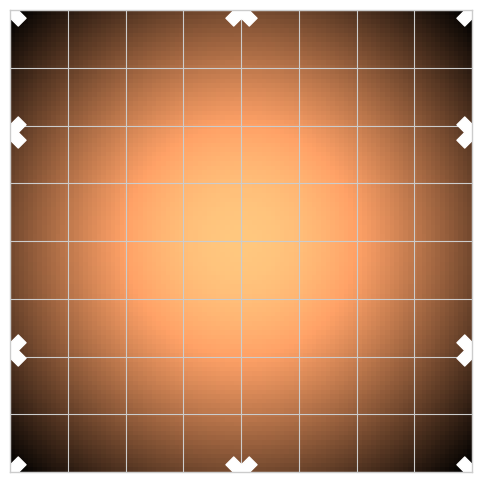} \\
\rotatebox[origin=c]{0}{qVS}
& \includegraphics[scale=0.18]{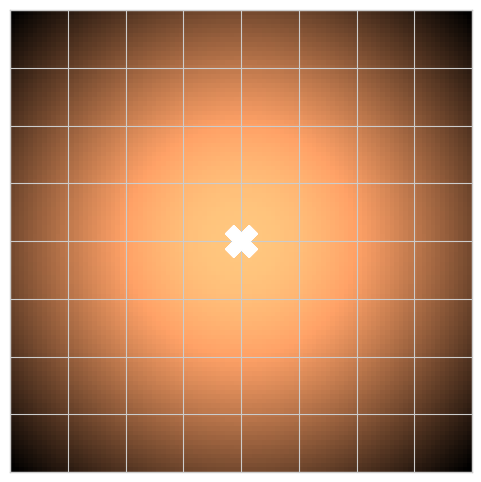}
& \includegraphics[scale=0.18]{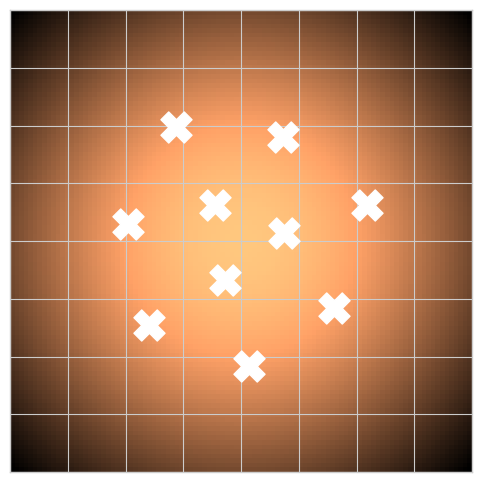}
& \includegraphics[scale=0.18]{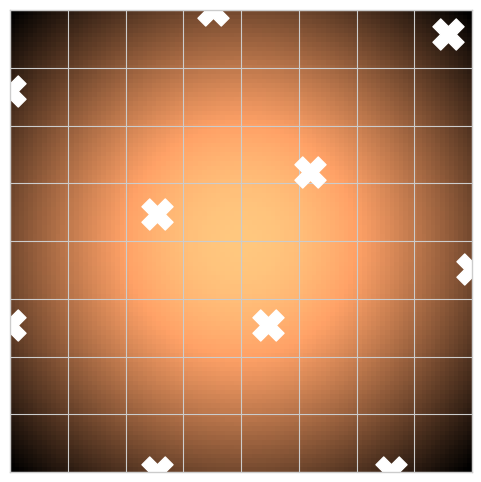}
& \includegraphics[scale=0.18]{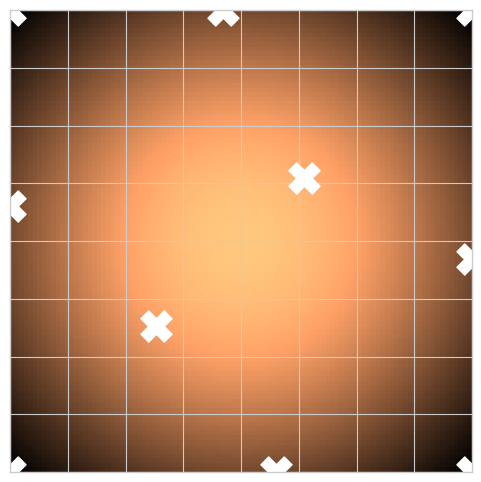}
& \includegraphics[scale=0.18]{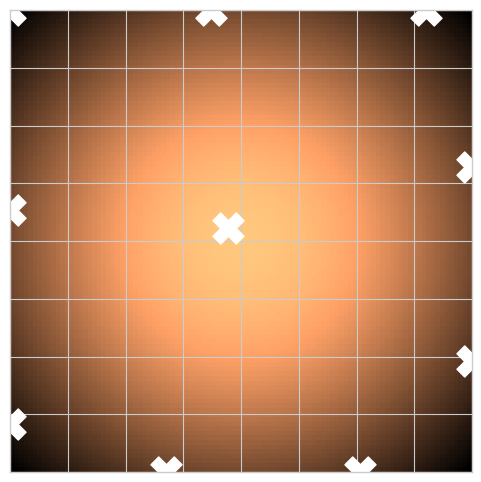} \\
& \multicolumn{1}{c}{$q = 0$}
& \multicolumn{1}{c}{$q = 0.1$}
& \multicolumn{1}{c}{$q = 1$}
& \multicolumn{1}{c}{$q = 5$}
& \multicolumn{1}{c}{$q = \infty$} \\
\end{tabularx}
\caption{
Batches of 10 data points maximizing various VS and qVS functions, obtained with multi-start gradient-based optimization.
The scoring function is a Gaussian function centered at the middle point, as illustrated by the heat maps.
The quality-weighted Vendi Score balances between the quality of the selected data points and their diversity; this balance is smoothly controlled by the order $q$.
}
\label{fig:toy}
\end{figure*}

Many real-world tasks can be framed as expensive discovery problems, where one explores large databases in search of rare, valuable items.
For instance, a scientist aiming to find a drug for a disease may need to iterate over millions of molecules to discover those that bind to specific biological targets. These search problems also often involve an expensive labeling process: the scientist will need to perform costly, time-consuming experiments to test for a molecule's binding activity to study its characteristics. This high cost in experimentation rules out exhaustive search and motivates the need for more sophisticated search strategies. 

Powerful active search (AS) and Bayesian optimization (BayesOpt) techniques have been developed over the years~\citep{garnett2012bayesian,jiang2017efficient,nguyen2021nonmyopic, nguyen2023nonmyopic, eriksson2019scalable} to tackle the problems mentioned above. While these advances have led to a more flexible experimental design framework, there has been a recent surge in interest in modifying existing algorithms to not only perform effective search/optimization but also induce more \emph{diversity} during experimentation.
For example, \citet{malkomes2021beyond} proposed an AS method in which data points sufficiently close to discoveries already made are removed from the pool, effectively encouraging a more diverse search.
\citet{maus2023discovering}, on the other hand, formulated a BayesOpt problem where the goal is to maintain and optimize multiple solutions that are constrained to be different from one another. These works leverage local penalization to induce diversity. 

Even though local penalization is a natural way to encourage diversity, it requires defining constraints to control the algorithms' behavior. However, setting these constraints can be challenging for complex, high-dimensional spaces. In this work, we present an alternative approach to local penalization to enforce diversity in experimental design. More specifically, we extend the Vendi scores (VS) \citep{friedman2023vendi, pasarkar2023cousins} to account for the quality of the items in a given input set. We call these new scores \emph{quality-weighted Vendi scores}.

These quality-weighted Vendi scores offer us a mathematically convenient way to evaluate quality and diversity and yield a unified framework for diverse experimental design. 
We applied the quality-weighted Vendi scores to both AS and BayesOpt across a wide variety of tasks involving drug discovery, materials discovery, and reinforcement learning. For all these tasks, we compared against strong existing baselines for experimental design, as well as analyzed the performance of our algorithms for various quality-diversity trade-offs. In all our experiments, we found that experimental design algorithms leveraging quality-weighted Vendi scores tend to outperform their counterparts in terms of both the diversity and the quality of the data points they yield.

\section{The Quality-Weighted Vendi Score}
\label{sec:method}

We first provide background on the Vendi score (VS) as a metric of diversity of a given set and introduce our extension to the VS that incorporates quality scores of individual members.
We then examine computationally efficient ways to optimize this quality-weighted VS, which will allow us to tackle a wide range of experimental design tasks in Section \ref{sec:exp_design}.

\subsection{The Vendi Score}

Consider a finite set of data points $X = \{ x_i \}_{i = 1}^n$ in some domain $\mathcal{X}$. \citet{friedman2023vendi} introduced the Vendi Score (VS) to characterize the diversity of a collection of items such as $X$. VS is defined as the exponential of the Shannon entropy of the normalized eigenvalues of the kernel similarity matrix corresponding to $X$. Specifically, given a positive semidefinite similarity function $k: \mathcal{X} \times \mathcal{X} \rightarrow \mathbb{R}$, where $k(x, x) = 1$ for all $x \in \mathcal{X}$, denote by $K \in \mathbb{R}^{n \times n}$ the kernel matrix corresponding to the set $X = \{ x_i \}_{i = 1}^n$ where each entry $K_{i, j} = k(x_i, x_j)$.
Further denote the eigenvalues of $K$ as $\lambda_1, \lambda_2, \ldots, \lambda_n$.
The VS is defined as:
\begin{equation}
\label{eq:vs}
\mathrm{VS}(X; k) = \exp \left( - \sum_{i = 1}^n \overline{\lambda}_i \, \log \overline{\lambda}_i \right),
\end{equation}
where $\overline{\lambda}_1, \overline{\lambda}_2, \ldots, \overline{\lambda}_n$ are the \emph{normalized} eigenvalues of $K$ such that $\overline{\lambda}_i = \lambda_i / \sum_{i = 1}^n \lambda_i$, and $0 \, \log 0$ is defined to be $0$.
As $K$ is a positive semidefinite matrix, the eigenvalues $\lambda_1, \lambda_2, \ldots, \lambda_n$ are non-negative and the normalized eigenvalues $\overline{\lambda}_1, \overline{\lambda}_2, \ldots, \overline{\lambda}_n$ sum to $1$.
The VS is then valid and can be viewed as the Shannon entropy of these normalized eigenvalues.

\citet{friedman2023vendi} explored the features of the VS as a diversity metric, and demonstrated that $\mathrm{VS}(X; k)$ can be interpreted as the effective number of unique samples of $X$.
In the extreme case where all items in $X$ are unique, $K$ is the identity matrix and $\mathrm{VS}(X; K) = n$. At the other end of the spectrum, if all items are identical, $K$ is the all-one matrix and $\mathrm{VS}(X; K) = 1$.
Overall, the VS offers us a mathematically principled yet convenient way to quantify the diversity of the items in a set $X$.
Unlike the determinantal point process (DPP) likelihoods \citep{kulesza2012determinantal}, another tool commonly used in diversity-related machine learning tasks, 
VS does not reduce to $0$ when there are duplicates in the input set $X$.
Compared to Hill numbers \citep{hill1973diversity}, the VS is not restricted to the assumption that different data points (species) are completely dissimilar to one another and thus allows us to effectively account for similarity between pairs of items.

\subsection{Accounting for Quality in the Vendi Score}

The VS captures diversity but treats all items in $X$ the same when it comes to their quality. In many situations, however, we may reasonably want our diversity metric to further express preference for items that exhibit desirable characteristics by incorporating ``quality scores'', upweighting or downweighting the final output based on the quality of individual items.
In other words, given a score function $s: \mathcal{X} \rightarrow \mathbb{R}$ that quantifies the quality of a given item $x \in \mathcal{X}$, we are interested in an extension to the VS that increases not only with more diverse but also with higher-quality items.

One may be tempted to mimic the DPPs' ability to naturally incorporate ``quality scores'' into their likelihoods, whereby the input kernel matrix is modified so that each entry $K_{i, j}$ is multiplied with the two corresponding quality scores $s(x_i)$ and $s(x_j)$.
Unfortunately, applying the same modification here for the VS does not lead to desirable results. The mismatch in behavior stems from the inherent mathematical difference between the two operations. The DPP likelihood can be interpreted as the volume spanned by the quality-weighted feature vectors of individual items $x_i$ whose outer products yield the kernel matrix. It therefore increases with higher values of $s(x_i)$.
The VS, on the other hand, computes the Shannon entropy of the normalized eigenvalues of the kernel matrix, which does not behave monotonically with respect to the values of the entries in $K$.

Our chosen solution to extend the VS to account for quality is simple---multiply the VS as defined in Eq. \ref{eq:vs} by the average quality score of individual items:
\begin{equation}
\label{eq:qvs}
\mathrm{qVS}(X; k, s) = \left( \sum_{i = 1}^n s(x_i) / n \right) \, \mathrm{VS}(X; k).
\end{equation}
This quality-weighted VS, qVS for short, possesses multiple desiderata partially inherited from the VS.
First, the output is maximized when all items are maximally diverse (the covariances $K_{i, j} = 0$) and achieve the highest quality score. Conversely, the qVS is minimized when all items are identical and yield the lowest quality score.
Fixing the quality scores $s(x_i)$, more diverse items (as measured by the VS) yield higher qVS values. Fixing the diversity score measured by the VS, higher-quality items yield higher qVS values. The intuitive interpretation of the VS as the effective number of unique samples carries over as well. The qVS can be interpreted as the effective number of high-quality samples in the input set. Overall, the qVS allows us to express our preference for diverse sets of high-quality data points.

\subsection{Controlling the Quality--Diversity Trade-off}

The balance between diversity and quality for a given set of items is explicitly achieved by maximizing the qVS in Eq. \ref{eq:qvs}.
However, in many scenarios, we may reasonably seek to control this balance to favor more diverse or higher-quality items, depending on our goal.
Inspired by the R\'enyi entropy, \citet{pasarkar2023cousins} proposed a generalization of the VS by introducing an extra hyperparameter $q \geq 0$, defining the VS of order $q$ as:
\begin{equation}
\label{eq:vs_q}
\mathrm{VS}_q (X; k) = \exp \left( \frac{1}{1 - q} \, \log \biggl( \sum_{i = 1}^n (\overline{\lambda}_i)^q \biggr) \right),
\end{equation}
where $\overline{\lambda}_1, \overline{\lambda}_2, \ldots, \overline{\lambda}_n$ are, again, the normalized eigenvalues of the kernel matrix $K$ corresponding to the input set $X$.
Further, when $q \in \{ 0, 1, \infty \}$, the $\mathrm{VS}_q$ is defined as the limit of Eq. \ref{eq:vs_q} as $q$ approaches the target order.
We briefly note that when $q = 1$, we recover the traditional VS that is the Shannon entropy of the normalized eigenvalues of the similarity matrix.

The order $q$ smoothly controls the sensitivity to the non-uniformity of these eigenvalues, and thus the evaluation of the diversity of $X$.
A smaller value of $q$ leads to a VS that is more sensitive to $X$ in that the VS increases faster than that of a larger $q$ when we add items to $X$.
In the extreme case, $\mathrm{VS}_0$ is simply the count function, which increases by $1$ every time a new data point is added to $X$, ignoring the diversity of the set.
For a larger $q$, it takes a completely unique data point (with zero covariances with other items) to lead to an increase of $1$ in the VS.
We use this more general VS in Eq. \ref{eq:qvs} to effectively balance quality and diversity. 

\subsection{Optimizing the Quality-Weighted Vendi Score}
\label{sec:qvs_opt}

With the qVS in hand, we can evaluate the value of a given input set $X$, assessing its diversity and the quality of its members.
A question naturally arises: given the domain $\mathcal{X}$, how can we identify a subset of a particular size that maximizes the qVS?
That is, we want to find:
\begin{equation}
\label{eq:max_qvs}
X_* = \argmax\limits_{X \subset \mathcal{X}, \mid X \mid = n} \mathrm{qVS}(X; k, s).
\end{equation}
As the VS in Eq. \ref{eq:vs} is differentiable, if the domain $\mathcal{X}$ is a compact space, the qVS can be efficiently maximized using an off-the-shelf gradient-based optimizer with multi-start, assuming that we also have access to the gradients of the kernel $k$ as well as the score function $s$.

The panels of Fig. \ref{fig:toy} show an example in 2 dimensions within $\mathcal{X} = [-1, 1]^2$.
Here, we use an isotropic Gaussian kernel: $k(\boldsymbol{x}_1, \boldsymbol{x}_2) = \exp \left( - \| \boldsymbol{x}_1 - \boldsymbol{x}_2 \|^2 / 2 \right)$, and a Gaussian score function centered at the origin: $s(\boldsymbol{x}) = \exp \left( - \| \boldsymbol{x} \|^2 / 2 \right)$, shown as the heat maps.
We run a gradient-based optimizer with multi-start to find a batch $X_*$ of size $n = 10$ maximizing the VS (top row) and the qVS (bottom row) of different orders $q$ and show the members of the optimal batches as red x's.
We notice interesting distinctions between the panels.
The data points at the top maximizing the VS are well spread out across the domain, maximizing pure diversity. Further, as $q$ increases, groups of close-by data points are discouraged, and the points are pushed towards the boundary. The data points maximizing the qVS at the bottom, on the other hand, favor points that yield high values for the score function while still achieving excellent diversity among the items. This figure showcases the qVS' ability to account for both quality and diversity, with the order q serving to flexibly control the balance between the two, with priority for diversity growing as $q$ increases.

If we can only evaluate $k$ or $s$ in a black-box manner, or if the domain $\mathcal{X}$ is a discrete set of items, optimization of the qVS becomes more challenging.
In fact, given a discrete search space $\mathcal{X}$, exact maximization of Eq. \ref{eq:max_qvs} is a combinatorial problem, as we need to iterate over all possible subsets $X$ of size $n$ in search for the optimal $X_*$.
We instead opt for approximate maximization of the qVS using the sequential greedy heuristic \citep{nemhauser1978analysis,krause2007near}, which has found application in maximizing functions with diminishing returns, including DPPs' likelihoods.
More specifically, we sequentially build the approximately optimal batch $\overline{X}_*$ from the empty set, finding the item $x$ that yields the largest increase in the qVS at each iteration:
\begin{equation}
x_* = \argmax_{x \in \mathcal{X} \setminus \overline{X}_*} \, \mathrm{qVS} \left( \overline{X}_* \cup \{ x \}; k, s \right) - \mathrm{qVS} \left( \overline{X}_*; k, s \right).
\end{equation}

We repeat this greedy search until the running batch $\overline{X}_*$ reaches the desirable size, at which point we return $\overline{X}_*$ as the set that approximately maximizes the qVS.
While we do not prove any theoretical guarantee in terms of optimization performance of the greedy strategy on the qVS, we empirically observe that the procedure works well in our experiments and is efficient enough to run at scale.

\section{Designing Diverse Experiments with the Vendi Score}
\label{sec:exp_design}

We are tasked with sequentially querying an expensive-to-evaluate oracle to obtain observations of a system of interest.
At each iteration of this procedure, we train a probabilistic model on the data collected so far and use the predictions of this model to decide which data point to label next.
This process is repeated for a pre-specified number of iterations.
The goal is to design effective ways to collect more data to maximize a metric of interest at the end of the procedure.

\subsection{Diverse Active Search}

Given a large but finite pool of unlabeled data $\mathcal{X}$, we seek to identify data points belonging to a rare, valuable class of interest.
We label these valuable data points, referred to as \emph{positives}, with $y = 1$ and use $y = 0$ for the other points, referred to as \emph{negatives}.
The label of a data point is not known \emph{a priori}, but can be determined by querying an expensive oracle.
Traditional AS targets achieving the highest ``hit rate'', that is, maximizing the number of positives in the collected data, $\sum_{(x, y) \in \mathcal{D}} \, y$,
where $\mathcal{D}$ is the collection of data we have chosen to label at the end of the search.

Given this formulation, active search (AS) strategies tend to become too exploitative, making many observations within regions in the search space $\mathcal{X}$ known to yield a high hit rate.
As \citet{nguyen2023nonmyopic} argued, in settings such as scientific discovery, there are diminishing returns in making additional discoveries in a frequently observed region: ``a discovery  in a novel region of the design space may offer more marginal insight than the 100th discovery in an already densely labeled region.''
We thus aim at an alternative AS setting that rewards diverse discoveries.
As the VS has been established as a principled diversity metric, we propose to directly use it to measure our search performance when diversity is of interest and modify the AS objective to be maximized to be:
\begin{equation}
\label{eq:as_obj}
\mathrm{VS_q} \big( \mathcal{D}_+; k \big),
\end{equation}
where the operator $_+$ gives the subset of positives within a set: 
\begin{equation}
\mathcal{D}_+ \triangleq \{ x \mid (x, y) \in \mathcal{D}, y = 1 \}.
\end{equation}
Our goal in this diversity-aware search is to collect a set of diverse positives.
Interestingly, setting $q = 0$ recovers the base version of AS, where our utility function is the count function.
We thus view our formulation as a generalization of traditional AS.

How should we design our queries to the oracle, sequentially selecting among the unlabeled data, so as to maximize the objective defined above?
Assuming access to a probabilistic predictive model that outputs, $\Pr \left( y = 1 \mid x, \mathcal{D} \right)$, the probability that an unlabeled item $x$ has a positive label given the data observed so far, we can derive the one-step Bayesian optimal decision, the data point $x_*$ that maximizes the expected increase of the objective in Eq. \ref{eq:as_obj}:
\begin{equation}\label{eq:as_acq}
x_* = \argmax_{x \in \mathcal{X} \setminus \mathcal{D}} \quad
\mathbb{E} \left[ \mathrm{VS_q} \big( (\mathcal{D} \cup \{ x \})_+; k \big) \right]
- \mathrm{VS_q} \big( \mathcal{D}_+; k \big),
\end{equation}
where the expectation is taken with respect to the label of each unlabeled candidate $x \in \mathcal{X} \setminus \mathcal{D}$.

In a purely sequential regime where queries are made one after another, Bayesian decision theory guides us to select $x_*$ to greedily maximize our VS objective.
In batch settings where multiple queries are made simultaneously to maximize experimental throughput---which are common in the real world---one may be tempted to simply extend the search criterion in Eq. \ref{eq:as_acq} from a single candidate $x \in \mathcal{X} \setminus \mathcal{D}$ to a batch of queries $X \subset \mathcal{X} \setminus \mathcal{D}$, seeking to maximize the expected increase of the VS of the positives in $\mathcal{D} \cup X$.
This is a daunting task.
First, for any candidate batch $X$ of size $b$, we need to iterate over $2^b$ possible label combinations of this batch to compute the expected VS after making these queries.
Second, similar to the task of finding a batch to optimize the VS or the qVS, finding a batch to optimize the expected VS means searching over a combinatorial space, requiring exponential computational effort.

\begin{algorithm}[tb]
   \caption{qVS-AS for diverse active search}
   \label{alg:qvs_as}
\begin{algorithmic}[1]
    \State {\bfseries inputs} observations $\mathcal{D}$, query batch size $n$
    \State {\bfseries returns} query batch $X$ of size $n$ maximizing \ref{eq:as_criterion}
    \State $X \leftarrow \emptyset$ \Comment{sequentially built from the empty set}
    \For{$i \leftarrow 1, \ldots, n$}
        \For{$x \in \mathcal{X} \setminus \left( \mathcal{D} \cup X \right)$}
            \State $\alpha(x) = \mathrm{qVS} \Big( X_+(\mathcal{D}) \cup X \cup \{ x \}; k, p \Big)$ \Comment{candidate scored by the qVS}
        \EndFor
        \State $X \leftarrow X \cup \{ \argmax_{x \in \mathcal{X} \setminus \left( \mathcal{D} \cup X \right)} \alpha(x) \}$ \Comment{add candidate yielding largest qVS}
    \EndFor
\end{algorithmic}
\end{algorithm}

The difficulties above motivate us to find a computationally tractable alternative criterion to select the next queries at each iteration of a batch AS problem.
An ideal batch of queries should balance between high probability for the candidates to have positive labels (after all, positives are the target of our search) and diversity among both those queries and the already observed positives.
As its goal is precisely offering an evaluation function that balances between some metric of quality and diversity, we turn to the qVS for this task.
Specifically, we set the scoring function $s(x)$ to be exactly the probability of being a positive $p(x)$, and use the qVS of the positives within $\mathcal{D} \cup X$ as our diverse search criterion.
In other words, we seek to find:
\begin{align}
\label{eq:as_criterion}
\begin{split}
X_* & = \argmax_{X \subset \mathcal{X} \setminus \mathcal{D}} \, \mathrm{qVS} \big( \mathcal{D}_+ \cup X; k, p \big) \\
& = \argmax_{X \subset \mathcal{X} \setminus \mathcal{D}} \left( \sum_{x \in X_+(\mathcal{D}) \cup X} \Pr \left( y = 1 \mid x, \mathcal{D} \right) / n \right) \mathrm{VS} \big( \mathcal{D}_+ \cup X; k \big),
\end{split}
\end{align}
where $n = \lvert \mathcal{D}_+ \cup X \rvert$.
Our last step is to identify the batch $X$ that maximizes this qVS metric, and we appeal to the methods described in \ref{sec:qvs_opt} for this task.
We show the pseudocode for our algorithm in Algorithm \ref{alg:qvs_as}.

\subsection{Diverse Bayesian Optimization}

Bayesian optimization (BayesOpt) \citep{garnett2022bayesian} is a framework for optimizing black-box functions.
Given a domain $\mathcal{X}$, which can be either discrete or continuous, BayesOpt sets out to find the global optimum of an objective function $f$ of interest: 
\begin{align*}
    x_* &= \argmax_{x \in \mathcal{D}} f(x)
    .
\end{align*}
Unlike active search (AS) where we work with binary labels, BayesOpt deals with real-valued labels $y$ that are the outputs of the objective function $f$.
Further, while AS focuses on finding many valuable data points within a search space, BayesOpt targets the singular, most valuable data point, tackling a different yet also relevant class of discovery problems commonly encountered in experimental design.

Similar to our discussion on AS, the pure-optimization formulation of BayesOpt often leads to overly exploitative strategies.
Even if a BayesOpt algorithm can effectively identify the global optimum of the objective function, \citet{maus2023discovering} argued this ``all-or-nothing'' goal of finding a single best solution to a problem is undesirable in many scenarios.
For example, when trying to discover metal-organic frameworks (MOFs) with high capacity for storage of toxic gasses, scientists may apply BayesOpt to identify a hypothetical MOF that, when simulated by a computer program, possesses a high storage capacity.
However, this high-performing MOF may turn out to be infeasible to synthesize in practice due to having unrealistic physical attributes. This leads to wasted resources and efforts. Having a diverse set of MOF candidates would give the scientist higher chances at finding a synthesizable MOF that meats the target criteria.

\citet{maus2023discovering} proposed a modified formulation that aims to find many diverse solutions.
Their framework involves maintaining a set of possible solutions that are of high quality \emph{and} diverse, where diversity is enforced by constraining the solutions to be at least a pre-specified distance away from one another.
Formally, denote by $\delta$ a distance function for an objective function $f$ of interest, they seek a sequence of $M$ solutions $\{ x^*_1, x^*_2, \ldots, x^*_M \}$ such that:
\begin{align}
\label{eq:robot}
\begin{split}
x^*_1 & = \argmax_{x \in \mathcal{X}} f(x), \\
x^*_i & = \argmax_{x \in \mathcal{X}} f(x) \text{ subject to } \delta(x^*_i, x^*_j) \geq \tau, \forall j < i,
\end{split}
\end{align}
where $\tau$ is a user-specified distance threshold that controls the diversity of the resulting solutions.
\citet{maus2023discovering} dubbed this formulation \emph{rank-ordered} BayesOpt, as the solutions $x^*_1, x^*_2, \ldots, x^*_M$ are ranked in that each subsequent solution is constrained to be far away from those that precede it.
The authors further proposed extending the trust region-based BayesOpt algorithm TuRBO \citep{eriksson2019scalable} to this setting.
TuRBO tackles high-dimensional problems via local optimization and consists of a set of local optimizers.
Each optimizer maintains a trust region around a promising region, and the size of the region expands or shrinks based on optimization performance.
(Local optimization is often accomplished with Thompson sampling \citep{russo2018tutorial} within each trust region.)
This strategy is particularly amenable to the rank-ordered formulation above, as one could center a trust region around each member of the solution set $x^*_i$, and iteratively refine that member via local optimization while obeying the diversity constraints.
The resulting algorithm, called ROBOT, was shown to be able to identify diverse and high-quality solutions in several tasks.

\begin{algorithm}[tb]
   \caption{qVS-BayesOpt with TuRBO for diverse Bayesian optimization}
   \label{alg:qvs_bayesopt_turbo}
\begin{algorithmic}[1]
    \State {\bfseries inputs} observations $\mathcal{D}$, number of trust regions $M$, query batch size $n$
    \State {\bfseries returns} query batch $X$ of size $n$ maximizing \ref{eq:qvs_robot}
    \For{$m \leftarrow 1, \ldots, M$} \Comment{generate candidates in each trust region}
        \State $\overline{X}_m = \mathrm{TuRBO}_m(\mathcal{D})$
    \EndFor

    \State $\overline{X} \leftarrow \cup_{m = 1}^M \overline{X}_m$ \Comment{merge all candidates}
    \State $X \leftarrow \emptyset$ \Comment{sequentially built from the empty set}
    
    \For{$i \leftarrow 1, \ldots, n$}
        \For{$x \in \overline{X} \setminus X$}
            \State $\alpha(x) = \mathrm{qVS} \Big( X \cup \{ x \}; k, \bar{f} \Big)$ \Comment{candidate scored by the qVS}
        \EndFor
        \State $X \leftarrow X \cup \{ \argmax_{x \in \overline{X} \setminus X} \alpha(x) \}$ \Comment{add candidate yielding largest qVS}
    \EndFor
\end{algorithmic}
\end{algorithm}

As mentioned, the goal of rank-ordered BayesOpt is to collect diverse data points that yield high objective values.
We propose to also use our qVS for this task and seek:
\begin{align}
\label{eq:qvs_robot}
\begin{split}
X_* & = \argmax_{X \subset \mathcal{X}} \, \mathrm{qVS_q} (X; k_\delta, f) \\
& = \argmax_{X \subset \mathcal{X}} \, \Big( \sum_{x \in X} f(x) / M \Big) \, \mathrm{VS_q} (X; k_\delta),
\end{split}
\end{align}
where $M = \lvert X \rvert$ is the number of solutions we wish to return to the user, and $k_\delta$ is the similarity function derived from the distance function $\delta$ (by, for example, inversing or subtracting from a maximum distance).
This formulation removes the hyperparameter $\tau$ in \ref{eq:robot} that constraints the solutions to be at least some distance away.
We view this as a desirable feature in many instances, for example if the task of setting $\tau$ is not straightforward, which is often the case in high dimensions where reasoning about distances becomes challenging.
By relying on the qVS to automatically balance diversity and quality among our solutions, we avoid this hyperparameter that might be difficult to tune.
Furthermore, our algorithm can also take advantage of the trust region-based strategy of TuRBO. Specifically, at each iteration of the BayesOpt loop, we use Thompson sampling to generate samples $\bar{f}$ of the Gaussian process fitted on the objective function $f$ within a local region of the domain, and use these samples $\bar{f}$ in lieu of the actual $f(x)$ values to maximize the criterion in Eq. \ref{eq:qvs_robot}.
The selected data points that maximize the criterion are chosen as our queries at the current iteration. We give the pseudocode for our algorithm in Algorithm \ref{alg:qvs_bayesopt_turbo}.

Lastly, we acknowledge two potential concerns.
First, if the geometry of the domain of the problem in question is well understood, and the user is confident the constraints in the formulation of ROBOT in Eq. \ref{eq:robot} are desirable, then that method should indeed be preferred to ours, as ROBOT specifically adheres to the constraints provided to it.
Second, while the qVS and VS metrics do have a hyperparameter of their own---the order $q$---which controls the sensitivity to the diversity of the items, we argue that the basic form with $q = 1$ serves as a good starting point, and observe good performance of $q = 1$ in our experiments.
Empirically, a user may tune $q$ during a validation step prior to running actual experiments by observing the induced behavior on a toy example and adjusting $q$ to match their preference.
 \glsresetall

\section{Related Work}
\label{sec:related}

The quality-weighted diversity metric described in this paper, qVS, directly extends the Vendi score \citep{friedman2023vendi}, which has found use in a wide range of applications~\citep{pasarkar2023vendi, berns2023towards, wu2023self, liu2024diversity}. The qVS is an alternative to the commonly used likelihoods of determinantal point processes (DPP), which can also account for the quality of a set of items but do not have a natural interpretation suitable for \emph{evaluating} diversity and quality.

We use the qVS to tackle two specific experimental design problems, active search (AS) and Bayesian optimization (BayesOpt), which commonly model discovery tasks in science and engineering. While \citet{garnett2012bayesian} originally formulated AS as maximizing the raw number of discovered targets, multiple subsequent works have extended the AS framework to settings where diversity is of concern. \citet{vanchinathan2015discovering} considered an Upper Confidence Bound-style algorithm \citep{auer2002using} with an extra priority for diversity. \citet{malkomes2021beyond} proposed maximizing a coverage objective, defined as the sum of the volumes of hyperspheres drawn around the targets discovered. This coverage metric encourages the queries to be far away from one another. \citet{nguyen2023nonmyopic} considered a multiclass setting and opted for a metric that rewards diversity in the labels. 

In BayesOpt, diversity has been artificially induced to aid optimization, commonly via a DPP \citep{wang2018active,nava2022diversified}. \citet{maus2023discovering}, on the other hand, directly targeted discovering diverse, high-quality solutions. As discussed in \ref{sec:exp_design}, they extended the state-of-the-art TuRBO algorithm by constraining the current solutions to be some distance away from one another. The resulting BayesOpt method ROBOT, as well as methods for AS mentioned earlier, crucially depend on accurately specifying the constraints to achieve the desired diversity. However, these distance constraints can be challenging to specify and enforce accurately, especially in high dimensions or with structured data. The qVS, in addition to being interpretable, offers a more flexible approach to diverse experimental design: instead of employing hard constraints on how far apart the solutions should be, we rely on the qVS to balance quality and diversity. The qVS, as we will show in \ref{sec:exp_design}, can be flexibly applied to many settings, including settings with non-continuous data for which TuRBO and ROBOT aren't applicable.

\begin{table*}[p!]
\centering
\caption{
Average Vendi Scores under different orders $q$ across 10 repeated experiments of the molecular discovery problem with the photoswitch data; the best performance in each column is highlighted in \textbf{bold} (including ties).
A star ($^*$) superscript indicates that the reported result is chosen from the best hyperparameter in that setting.
}
\label{tab:varying_q_photoswitch}
\resizebox{\textwidth}{!}{\begin{tabular}{clccccccccc}
\toprule
\multicolumn{2}{c}{\multirow{2}{*}{method}} & \multicolumn{6}{c}{Vendi Score} & \multirow{2}{*}{max.\ pairwise dist.\ } & \multirow{2}{*}{kernel matrix det.\ } \\
\cmidrule(lr){3-8}
& & $q = 0$ & $q = 0.1$ & $q = 0.5$ & $q = 1$ & $q = 2$ & $q = \infty$ \\
\midrule

\multicolumn{2}{c}{random search} & \resultalt{72.20}{1.96} & \resultalt{60.66}{1.78} & \resultalt{44.30}{1.34} & \resultalt{30.59}{0.98} & \resultalt{18.42}{0.70} & \resultalt{7.96}{0.43} & \resultalt{0.94}{0.00} & \resultalt{0.32}{0.01} \\

\multicolumn{2}{c}{ECI$^*$} & \resultalt{67.10}{1.93} & \resultalt{56.65}{1.24} & \resultalt{42.07}{0.70} & \resultalt{29.70}{0.35} & \resultalt{18.52}{0.34} & \resultalt{8.64}{0.26} & \resultboldalt{0.95}{0.00} & \resultalt{0.37}{0.02} \\

\multicolumn{2}{c}{SELECT $^*$} & \resultalt{125.90}{5.70} & \resultalt{97.33}{4.65} & \resultalt{58.07}{1.21} & \resultalt{30.09}{1.47} & \resultalt{12.53}{1.50} & \resultalt{4.40}{0.43} & \resultalt{0.95}{0.00} & \resultalt{0.06}{0.01} \\

\midrule
\multirow{6}{*}{qVS-AS} & $q = 0$ & \resultboldalt{167.50}{11.03} & \resultalt{105.70}{6.32} & \resultalt{52.78}{2.39} & \resultalt{20.83}{0.45} & \resultalt{7.30}{0.18} & \resultalt{3.28}{0.03} & \resultalt{0.93}{0.00} & \resultalt{0.02}{0.02} \\

& $q = 0.1$ & \resultalt{139.60}{4.76} & \resultboldalt{120.26}{3.80} & \resultboldalt{67.29}{1.55} & \resultalt{31.71}{0.82} & \resultalt{12.68}{0.51} & \resultalt{5.19}{0.32} & \resultalt{0.95}{0.00} & \resultalt{0.03}{0.01} \\

& $q = 0.5$ & \resultalt{72.00}{1.22} & \resultalt{67.91}{1.11} & \resultalt{53.53}{0.83} & \resultboldalt{39.85}{0.75} & \resultalt{24.55}{0.69} & \resultalt{9.01}{0.28} & \resultalt{0.95}{0.00} & \resultalt{0.39}{0.01} \\

& $q = 1$ & \resultalt{42.50}{0.49} & \resultalt{41.39}{0.51} & \resultalt{38.11}{0.46} & \resultalt{34.73}{0.41} & \resultboldalt{29.72}{0.36} & \resultalt{14.46}{0.39} & \resultalt{0.95}{0.00} & \resultalt{0.74}{0.00} \\

& $q = 2$ & \resultalt{33.70}{0.74} & \resultalt{33.39}{0.73} & \resultalt{32.25}{0.71} & \resultalt{30.98}{0.68} & \resultalt{28.88}{0.65} & \resultboldalt{17.37}{0.40} & \resultalt{0.94}{0.00} & \resultalt{0.85}{0.00} \\

& $q = \infty$ & \resultalt{21.90}{0.68} & \resultalt{21.84}{0.68} & \resultalt{21.59}{0.64} & \resultalt{21.29}{0.60} & \resultalt{20.75}{0.53} & \resultalt{15.93}{0.32} & \resultalt{0.94}{0.00} & \resultboldalt{0.93}{0.00} \\

\bottomrule
\end{tabular}
}

\vskip 50pt
\caption{
Average Vendi Scores under different orders $q$ across 10 repeated experiments of the materials discovery problem with the bulk metal glass data; the best performance in each column is highlighted in \textbf{bold} (including ties).
A star ($^*$) superscript indicates that the reported result is chosen from the best hyperparameter in that setting.
}
\label{tab:varying_q_bmg}
\resizebox{\textwidth}{!}{\begin{tabular}{clccccccccc}
\toprule
\multicolumn{2}{c}{\multirow{2}{*}{method}} & \multicolumn{6}{c}{Vendi Score} & \multirow{2}{*}{max.\ pairwise dist.\ } & \multirow{2}{*}{kernel matrix det.\ } \\
\cmidrule(lr){3-8}
& & $q = 0$ & $q = 0.1$ & $q = 0.5$ & $q = 1$ & $q = 2$ & $q = \infty$ \\
\midrule
\multicolumn{2}{c}{random search} & \resultalt{7.20}{0.86} & \resultalt{7.18}{0.85} & \resultalt{7.09}{0.82} & \resultalt{6.99}{0.79} & \resultalt{6.82}{0.74} & \resultalt{5.42}{0.41} & \resultalt{13.16}{0.62} & \resultalt{0.88}{0.08} \\
\multicolumn{2}{c}{ECI$^*$} & \resultalt{25.00}{0.00} & \resultalt{24.89}{0.00} & \resultalt{24.46}{0.00} & \resultalt{23.92}{0.00} & \resultalt{22.85}{0.00} & \resultalt{12.85}{0.00} & \resultalt{15.35}{0.52} & \resultalt{0.99}{0.00} \\
\multicolumn{2}{c}{SELECT $^*$} & \resultalt{143.90}{2.85} & \resultalt{130.63}{4.47} & \resultalt{99.59}{5.66} & \resultalt{69.57}{6.12} & \resultalt{36.19}{5.13} & \resultalt{9.67}{1.46} & \resultalt{12.42}{0.60} & \resultalt{0.00}{0.00} \\
\midrule
\multirow{6}{*}{qVS-AS} & $q = 0$ & \resultboldalt{186.00}{3.04} & \resultalt{146.35}{4.20} & \resultalt{69.49}{5.14} & \resultalt{26.42}{3.86} & \resultalt{8.66}{1.45} & \resultalt{3.31}{0.36} & \resultalt{6.45}{0.54} & \resultalt{0.00}{0.00} \\
& $q = 0.1$ & \resultalt{174.30}{3.59} & \resultboldalt{162.82}{4.05} & \resultalt{120.07}{5.23} & \resultalt{75.92}{5.11} & \resultalt{31.67}{3.07} & \resultalt{8.13}{0.75} & \resultalt{10.49}{0.36} & \resultalt{0.00}{0.00} \\
& $q = 0.5$ & \resultalt{160.40}{3.07} & \resultalt{156.68}{3.08} & \resultalt{141.15}{3.13} & \resultalt{120.74}{3.20} & \resultalt{83.15}{3.14} & \resultalt{20.94}{1.19} & \resultalt{15.08}{0.39} & \resultalt{0.00}{0.00} \\
& $q = 1$ & \resultalt{153.70}{1.24} & \resultalt{151.93}{1.23} & \resultboldalt{144.48}{1.19} & \resultboldalt{134.45}{1.17} & \resultalt{113.21}{1.24} & \resultalt{35.32}{1.28} & \resultalt{18.58}{0.70} & \resultalt{0.31}{0.05} \\
& $q = 2$ & \resultalt{137.60}{1.72} & \resultalt{136.82}{1.70} & \resultalt{133.59}{1.65} & \resultalt{129.32}{1.57} & \resultboldalt{120.19}{1.38} & \resultalt{50.55}{0.65} & \resultalt{22.34}{0.35} & \resultalt{0.96}{0.00} \\
& $q = \infty$ & \resultalt{65.40}{4.40} & \resultalt{65.34}{4.39} & \resultalt{65.11}{4.33} & \resultalt{64.82}{4.26} & \resultalt{64.25}{4.13} & \resultboldalt{52.71}{2.50} & \resultboldalt{25.58}{0.17} & \resultboldalt{1.00}{0.00} \\
\bottomrule
\end{tabular}
}

\vskip 50pt
\caption{
Average Vendi Scores under different orders $q$ across 10 repeated experiments of the product recommendation problem with the FashionMNIST data; the best performance in each column is highlighted in \textbf{bold} (including ties).
A star ($^*$) superscript indicates that the reported result is chosen from the best hyperparameter in that setting.
}
\label{tab:varying_q_fashion}
\resizebox{\textwidth}{!}{\begin{tabular}{clccccccccc}
\toprule
\multicolumn{2}{c}{\multirow{2}{*}{method}} & \multicolumn{6}{c}{Vendi Score} & \multirow{2}{*}{max.\ pairwise dist.\ } & \multirow{2}{*}{kernel matrix det.\ } \\
\cmidrule(lr){3-8}
& & $q = 0$ & $q = 0.1$ & $q = 0.5$ & $q = 1$ & $q = 2$ & $q = \infty$ \\
\midrule

\multicolumn{2}{c}{random search} & \resultalt{19.70}{1.64} & \resultalt{19.08}{1.45} & \resultalt{18.31}{1.30} & \resultalt{17.81}{1.20} & \resultalt{16.94}{1.02} & \resultalt{11.62}{1.20} & \resultalt{9.11}{1.07} & \resultboldalt{0.64}{0.10} \\

\multicolumn{2}{c}{ECI$^*$} & \resultalt{32.80}{1.39} & \resultalt{32.79}{1.39} & \resultalt{32.76}{1.38} & \resultalt{32.73}{1.37} & \resultalt{32.67}{1.35} & \resultboldalt{29.23}{1.34} & \resultalt{7.77}{1.03} & \resultalt{0.48}{0.08} \\

\multicolumn{2}{c}{SELECT $^*$} & \resultalt{135.40}{6.56} & \resultalt{65.89}{3.67} & \resultalt{45.65}{3.39} & \resultalt{31.88}{3.08} & \resultalt{20.17}{2.43} & \resultalt{7.59}{0.81} & \resultalt{1.79}{0.19} & \resultalt{0.00}{0.00} \\

\midrule
\multirow{6}{*}{qVS-AS} & $q = 0$ & \resultboldalt{155.00}{9.92} & \resultboldalt{69.90}{4.12} & \resultalt{40.81}{2.09} & \resultalt{21.92}{0.79} & \resultalt{10.58}{0.36} & \resultalt{4.39}{0.12} & \resultalt{1.45}{0.19} & \resultalt{0.00}{0.00} \\

& $q = 0.1$ & \resultalt{76.40}{4.65} & \resultalt{59.81}{3.54} & \resultalt{43.59}{2.69} & \resultalt{34.47}{2.25} & \resultalt{24.68}{1.87} & \resultalt{8.94}{0.64} & \resultalt{5.39}{0.60} & \resultalt{0.00}{0.00} \\

& $q = 0.5$ & \resultalt{65.40}{3.47} & \resultalt{60.67}{3.02} & \resultboldalt{54.88}{2.74} & \resultboldalt{49.92}{2.49} & \resultboldalt{43.80}{2.18} & \resultalt{22.34}{1.23} & \resultalt{7.90}{0.81} & \resultalt{0.00}{0.00} \\

& $q = 1$ & \resultalt{61.40}{3.77} & \resultalt{57.77}{3.36} & \resultalt{52.94}{3.11} & \resultalt{48.75}{2.86} & \resultalt{43.52}{2.53} & \resultalt{24.49}{1.25} & \resultalt{7.93}{0.86} & \resultalt{0.00}{0.00} \\

& $q = 2$ & \resultalt{46.70}{4.23} & \resultalt{44.09}{4.04} & \resultalt{41.13}{3.69} & \resultalt{38.47}{3.39} & \resultalt{35.03}{3.06} & \resultalt{21.75}{1.98} & \resultalt{8.32}{0.84} & \resultalt{0.04}{0.02} \\

& $q = \infty$ & \resultalt{25.50}{2.27} & \resultalt{25.29}{2.20} & \resultalt{24.51}{1.95} & \resultalt{23.71}{1.71} & \resultalt{22.59}{1.40} & \resultalt{19.50}{0.98} & \resultboldalt{9.75}{0.77} & \resultalt{0.50}{0.01} \\

\bottomrule
\end{tabular}
}
\end{table*}

\section{Experiments}
\label{sec:experiments}

We now present results from our numerical experiments, comparing active search (AS) and Bayesian optimization (BayesOpt) performance of our methods against a wide range of baselines.
In each experimental setting, we average results across 10 repeats with different initial data, chosen uniformly at random from the search space (these sets of 10 different initial data sets are shared across the methods).

\subsection{Diverse Active Search}
\label{sec:experiments_as}

We first discuss AS, where our goal is to collect a diverse set of positive points in a binary setting.
We study the performance of our method, which we call qVS-AS, under different orders $q \in \{ 0, 0.1, 0.5, 1, 2, \infty \}$, both in the search behavior as defined in \ref{eq:as_criterion} and in the evaluation metric.
Again note that $q = 0$ gives us the traditional AS setting \citep{garnett2012bayesian}, which counts the raw number of positives discovered and does not account for diversity.
We also consider relevant active learning/search algorithms discussed in \ref{sec:related}: Expected Coverage Improvement (ECI) \citep{malkomes2021beyond} and SELECT \citep{vanchinathan2015discovering}.
These methods come with their own hyperparameters to tune, and we only report the results obtained from the highest-performing hyperparameters.

We consider the molecular discovery problem studied by \citet{mukadum2021efficient}, where our target is photoswitches (molecules that change their properties upon irradiation) in chemical databases that exhibit both desirable light absorbance and long half-lives.
Roughly 36\% of the molecules in the search space are targets.
We also include a materials discovery application where we search for alloys that can form valuable bulk metallic glasses with higher toughness and better wear resistance than crystalline alloys. 
This data set comprises 106\,810 alloys from the materials literature, approximately 4\% of which exhibit glass-forming ability \citep{kawazoe1997nonequilibrium,ward2016general}.
Finally, following \citet{nguyen2023nonmyopic}, we use the FashionMNIST data set \citep{xiao2017fashion} of 70,000 images of articles of clothing to simulate a product recommendation problem.
Here, we assume that a user is looking for, unbeknownst to the recommendation engine, t-shirts and tops (members of class 0, one-tenth of the data set) while shopping online, and the goal is to assemble a diverse set of products belonging to this unknown class.

To first ensure the quality of our predictive model, we perform the following benchmarking experiments.
In each experiment, we train the model on 100 random points from a data set.
We then pick out the test points that yield the highest posterior probabilities $\Pr \left( y = 1 \mid x, \mathcal{D} \right)$ and record the proportion of this set are positives.
For each data set, we repeat this experiment 10 times and record the average precision-at-$k$, which ranges consistently from 80\% to 100\%, indicating that our model produces high-quality predictions and recovers pure sets of rare positives.

\begin{figure}
\centering
\includegraphics[width=\linewidth]{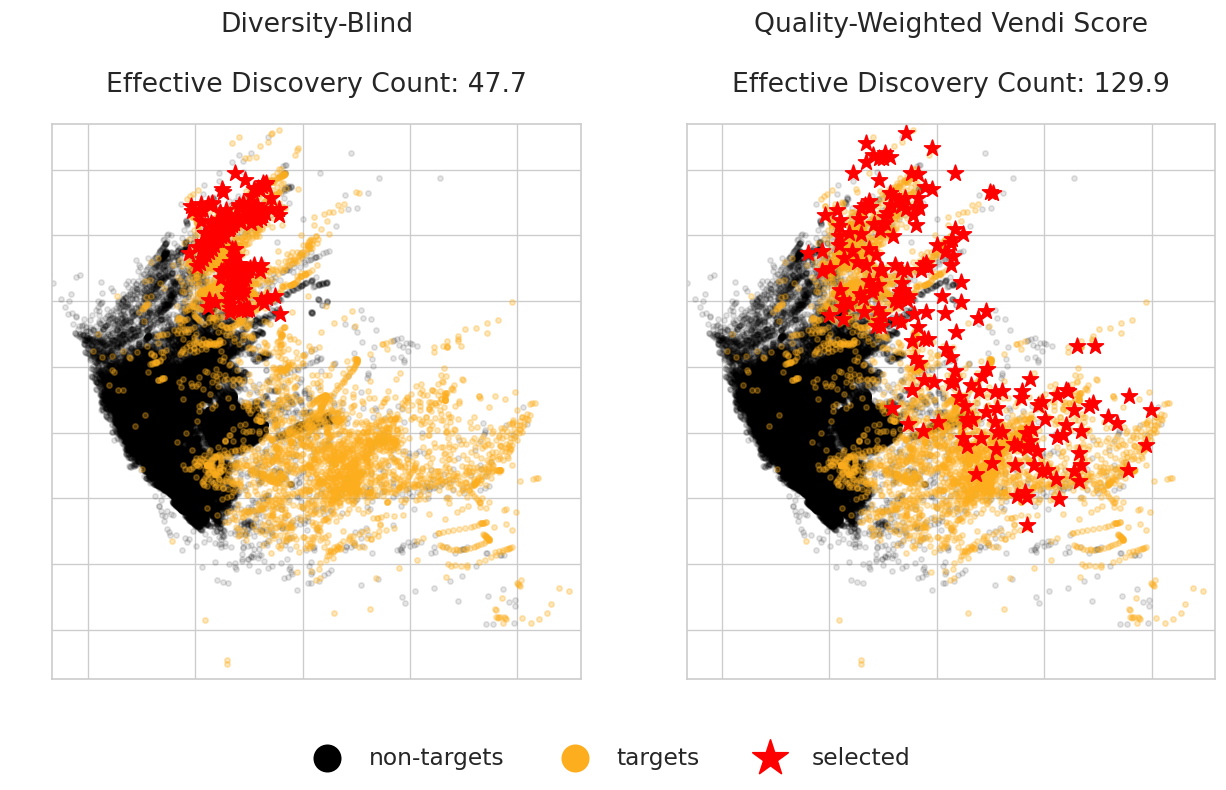}
\caption{
Data points collected by diversity-blind search and our diversity-aware policy in the materials discovery problem with bulk metal glasses.
Our method appropriately balances between exploring the search space and focusing on regions containing positive data, and discovers more effective positives as a result.
}
\label{fig:iter39}
\end{figure}

We use the VS of the collected positives in Eq. \ref{eq:as_obj} as our performance metric and show in the first portion of Tables \ref{tab:varying_q_photoswitch}, \ref{tab:varying_q_bmg}, and \ref{tab:varying_q_fashion} the VS (of different orders $q$) achieved by each method, averaged across the 10 repeats.
We see that our method qVS-AS performs well across the problems, achieving the highest VS in most cases; when it is not the best, it is typically a close second behind a method with tuned hyperparameters.
Inspecting the performance of different realizations of qVS-AS under varying values of the order $q$, we observe a reasonable trend: qVS-AS with the order $q$ matching that of the evaluation metric tends to perform the best; further, there is a smooth change in performance as we move across the different values of $q$, showcasing the ability of this hyperparameter to smoothly control our algorithm's behavior.

To further study the diversity of the data collected by each method, we consider two metrics that quantify the spread of the discovered targets: the maximum distance between any pair of positives discovered and the determinant of the kernel matrix of the collected positives (i.e., the squared volume spanned by the feature vectors of the selected data points with positive labels).
The last two columns in Tables \ref{tab:varying_q_photoswitch}, \ref{tab:varying_q_bmg}, and \ref{tab:varying_q_fashion}  show these results,
where our method can again be observed to achieve consistently good performance.

Finally, to visually illustrate our method's ability to assemble diverse data, we show in Fig. \ref{fig:iter39} the locations of the queries made by diversity-blind AS and our method, within the two-dimensional embedding of the bulk metal glass data set computed by performing PCA on the features.
We see that the overly exploitative diversity-blind search simply focuses on a small portion of the search space, while our method qVS-AS (with $q = 1$) is able to thorough explore the different regions of positives.
We also show the VS of the collected positives of the two policies (interpreted as the effective discovery count), where our policy clearly outperforms diversity-blind search.

\subsection{Diverse Bayesian Optimization}

\begin{figure*}[t]
\begin{subfigure}{0.33\textwidth}
    \includegraphics[width=\linewidth]{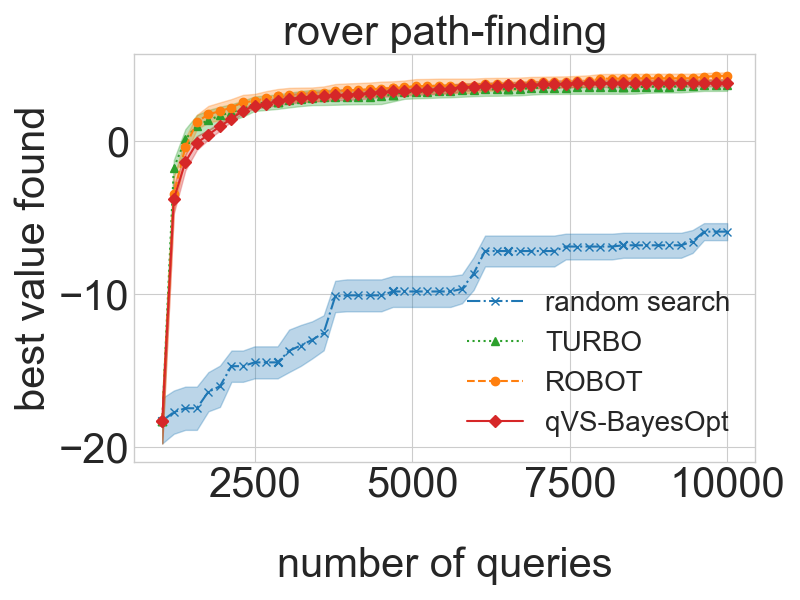}
\end{subfigure}\hfill
\begin{subfigure}{0.33\textwidth}
    \includegraphics[width=\linewidth]{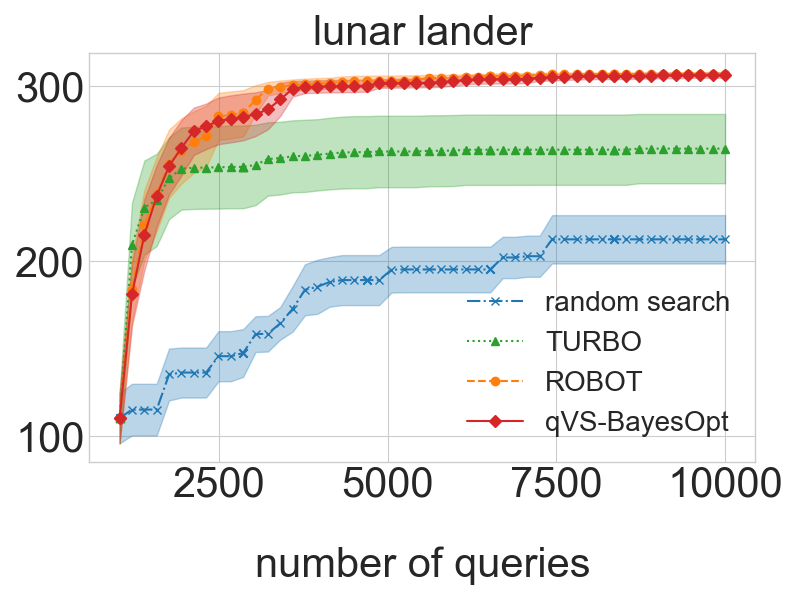}
\end{subfigure}\hfill
\begin{subfigure}{0.33\textwidth}
    \includegraphics[width=\linewidth]{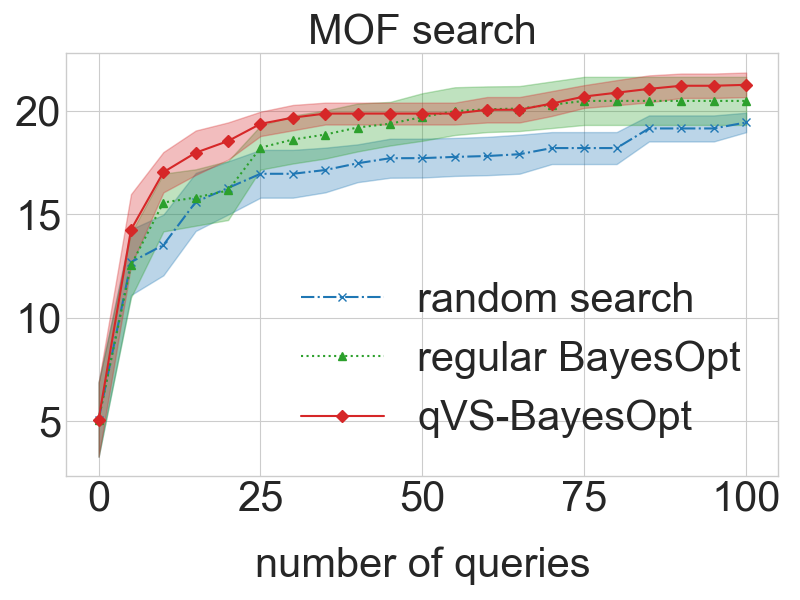}
\end{subfigure}
\caption{
Average optimization performance and standard errors across 10 repeated experiments.
Our method (shown in \textbf{\textcolor{red}{red}}) performs competitively across the different settings.
}
\label{fig:bayesopt_results}
\end{figure*}

\begin{figure*}[t]
\centering
\begin{subfigure}{0.4\textwidth}
    \includegraphics[width=\linewidth]{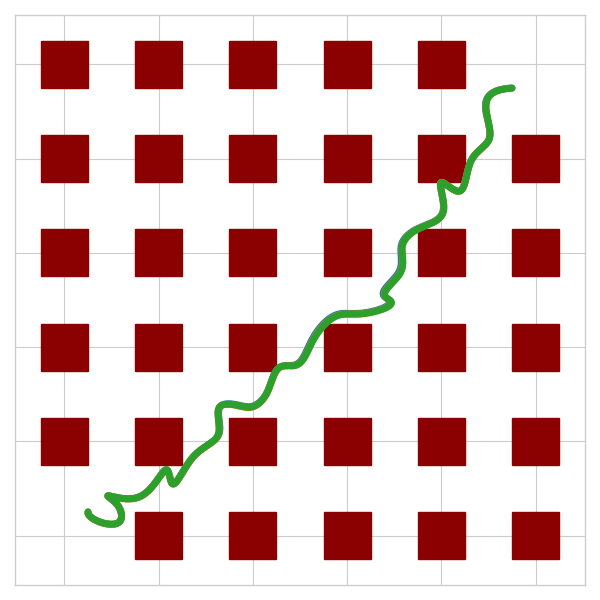}
    \subcaption{TuRBO (paths overlapping)}
\end{subfigure}\hskip 10pt
\begin{subfigure}{0.4\textwidth}
    \includegraphics[width=\linewidth]{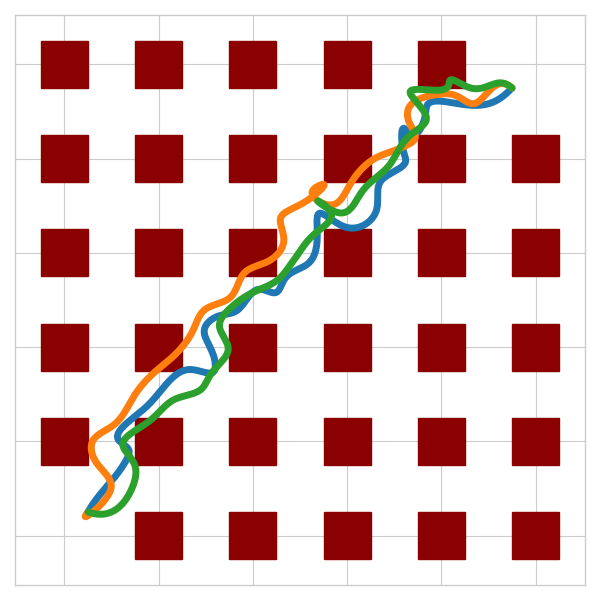}
    \subcaption{qVS-BayesOpt with $q = 0.5$}
\end{subfigure} \\
\begin{subfigure}{0.4\textwidth}
    \includegraphics[width=\linewidth]{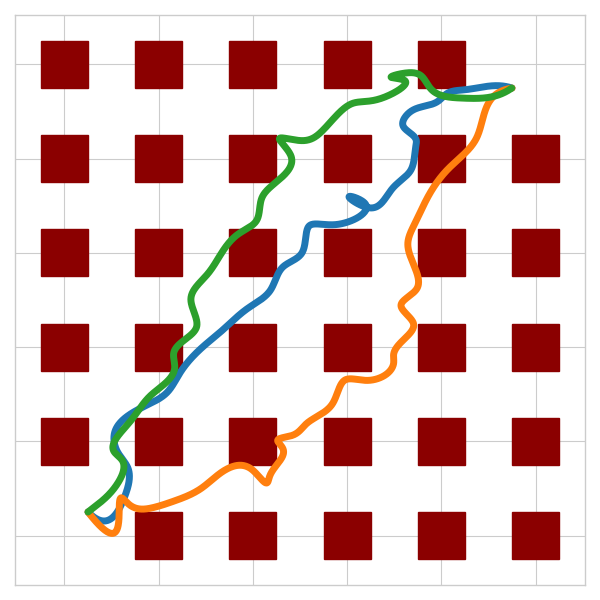}
    \subcaption{qVS-BayesOpt with $q = 1$}
\end{subfigure}\hskip 10pt
\begin{subfigure}{0.4\textwidth}
    \includegraphics[width=\linewidth]{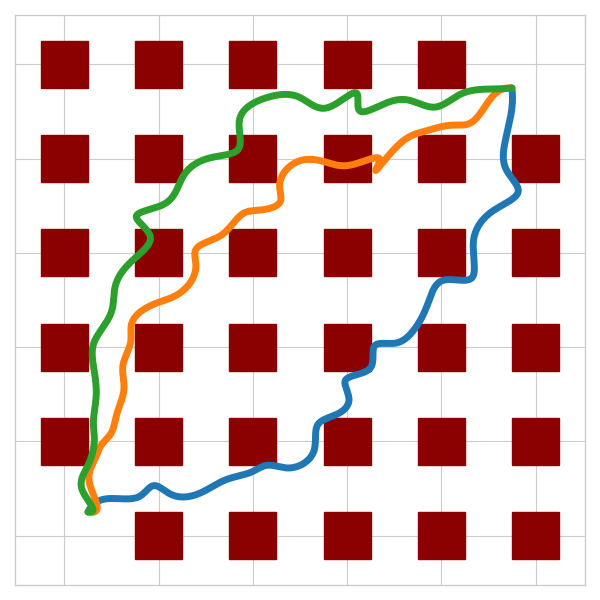}
    \subcaption{qVS-BayesOpt with $q = 1.5$}
\end{subfigure}
\vskip -5pt
\caption{
Trajectories identified by various search methods in the rover path finding problem.
Our method finds a diverse set of paths, whose diversity can be controlled using the order $q$ of the qVS.
}
\label{fig:rover}
\end{figure*}

We now present results from BayesOpt tasks, as formulated in Section \ref{sec:exp_design}.
To study the performance of our method, qVS-BayesOpt, we include as baselines (1) TuRBO \citep{eriksson2019scalable}, the diversity-blind algorithm upon which qVS-BayesOpt is based, (2) ROBOT \citep{maus2023discovering}, which also tackles diverse BayesOpt, and (3) a random search algorithm that uniformly samples its queries from the search space at random.
ROBOT has a hyperparameter $\tau$ that controls the quality--diversity trade-off in its search, which we set to the values used in the original investigation by \citet{maus2023discovering}.
We test these methods on three optimization tasks: (1) the rover path-finding task involves optimizing the path of a mars rover while avoiding obstacles; (2) the lunar lander task from reinforcement learning where we aim to optimize the control policy for an autonomous vehicle to safely land on a given terrain; (3) the metal-organic framework (MOF) storage capacity optimization task as explored by \citet{liu2024diversity}, where we aim to identify the MOFs that have the highest storage capacity for ammonia.

\begin{algorithm}[tb]
   \caption{qVS-BayesOpt for structured data within a discrete search space}
   \label{alg:qvs_bayesopt_discrete}
\begin{algorithmic}[1]
    \State {\bfseries inputs} observations $\mathcal{D}$, query batch size $n$
    \State {\bfseries returns} query batch $X$ of size $n$ maximizing \ref{eq:qvs_robot} in the discrete case
    \State $X \leftarrow \emptyset$ \Comment{sequentially built from the empty set}
    \For{$i \leftarrow 1, \ldots, n$}
        \For{$x \in \mathcal{X} \setminus \left( \mathcal{D} \cup X \right)$}
            \State $\alpha(x) = \mathrm{qVS} \Big( \mathcal{D} \cup X \cup \{ x \}; k, \mathrm{UCB} \Big)$ \Comment{candidate scored by the qVS}
        \EndFor
        \State $X \leftarrow X \cup \{ \argmax_{x \in \mathcal{X} \setminus \left( \mathcal{D} \cup X \right)} \alpha(x) \}$ \Comment{add candidate yielding largest qVS}
    \EndFor
\end{algorithmic}
\end{algorithm}

While the first two tasks are formulated as continuous optimization problems (60- and 12-dimensional, respectively), the third involves a discrete search space of structured data (a database of 1000 MOFs).
Typically, to deal with structured data such as molecules in BayesOpt, one may train a deep learning model such as a variational autoencoder (VAE) \citep{kingma2013auto} to obtain a continuous embedding of the candidates one searches over (see \citet{gomez2018automatic} for an example in drug discovery).
From there, one can apply BayesOpt algorithms such as TuRBO to that continuous embedding.
However, unlike drug-like molecules which have enjoyed enduring interest from the machine learning community, MOFs are relatively unexplored materials to which, to our knowledge, there does not exist any consistently suitable VAE that can be applied.
We instead reuse the MOF-specific kernel function proposed by \citet{liu2024diversity}, which operates on any given pair among the 1000 candidate MOFs.
Without a continuous embedding, the trust region-based algorithm TuRBO, and thus its extension to diverse BayesOpt, ROBOT, cannot be applied to this MOF search task.
Instead, we employ a simple Upper Confidence Bound (UCB) algorithm \citep{auer2002using} as our baseline of traditional BayesOpt.
To realize our algorithm with the qVS, we directly use the UCB score as our metric of quality in Eq. \ref{eq:qvs_robot} which we use as our criterion for finding the next queries (instead of the sample $\bar{f}$ from Thompson sampling in TuRBO).
Details of this algorithm is given in Algorithm \ref{alg:qvs_bayesopt_discrete}.

\begin{figure*}[t]
\centering
\includegraphics[width=\linewidth]{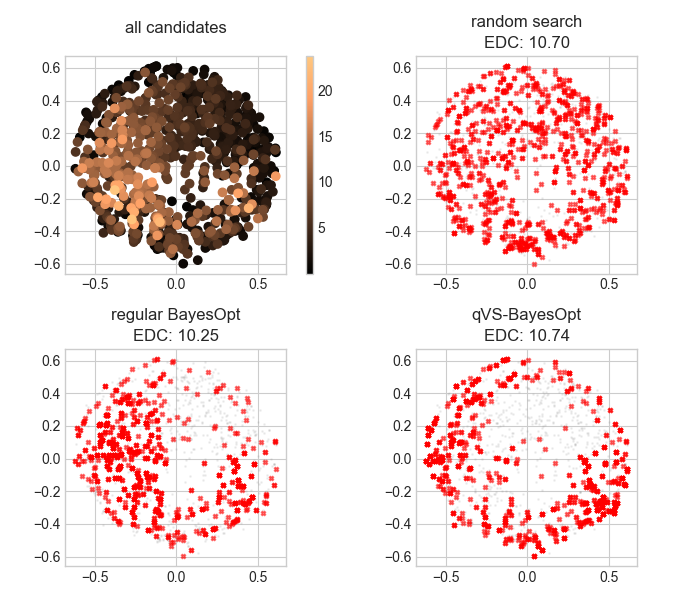}
\vskip -10pt
\caption{
Data points collected by each search strategy in the MOF search problem across the repeats, illustrated as \textbf{\textcolor{red}{red}} x's.
(All data points in the search space are visualized in the first panel.)
Our method focuses on specific, high-performing regions compared to the random search, while exploring the space more evenly compared to regular Bayesian optimization.
}
\label{fig:mof_scatter}
\end{figure*}

Results from these experiments are reported in Fig. \ref{fig:bayesopt_results}, where we show the highest objective value achieved across the 10 repeats.
We see that our method qVS-BayesOpt ($q = 1$) performs well across the experiments and remains competitive against the state-of-the-art ROBOT in the two continuous optimization problems.
Surprisingly, in the lunar lander and MOF search tasks, encouraging more diversity in our search not only does not result in any slowdown in optimization progress compared to traditional BayesOpt, but actually leads to improved performance.
We hypothesize this is because the search spaces in these two problems consist of many local optima in which exploitative BayesOpt algorithms could become trapped.
To highlight our method's ability to identify diverse solutions, we first show in Fig. \ref{fig:rover} the rover paths optimized by TuRBO, which targets pure optimization, in a representative run.
(Here, the number of solutions to be returned to the user $M = 3$.)
We see that these paths are effectively identical to and overlap one another.
On the other hand, the other panels show the set of 3 solutions optimized by qVS-BayesOpt from the run using the same initial data as TuRBO above, which exhibit varying degrees of diversity corresponding to $q \in \{ 0.5, 1, 1.5 \}$.

\begin{figure}
\parbox[t]{0.5\linewidth}{\null
\centering
\begin{tabular}{lc}
\toprule
& best storage value \\
\midrule
$q = 0$ & $20.47 ~ (1.16)$ \\
$q = 0.1$ & $22.07 ~ (0.81)$ \\
$q = 0.5$ & $20.07 ~ (0.61)$ \\
$q = 1$ & $21.24 ~ (0.59)$ \\
$q = 2$ & $20.23 ~ (0.56)$ \\
$q = \infty$ & $20.23 ~ (0.56)$ \\
\bottomrule
\end{tabular}
\vskip 18.5pt
\captionof{table}[t]{
Average storage capacity values (higher is better) and standard errors of the best MOFs found by our algorithm under different orders $q$.
Here, $q = 0$ corresponds to regular, diversity-blind Bayesian optimization.
}
\label{tab:mof_varying_q}
}
\hfill
\parbox[t]{0.45\linewidth}{\null
\centering
\includegraphics[width=\linewidth]{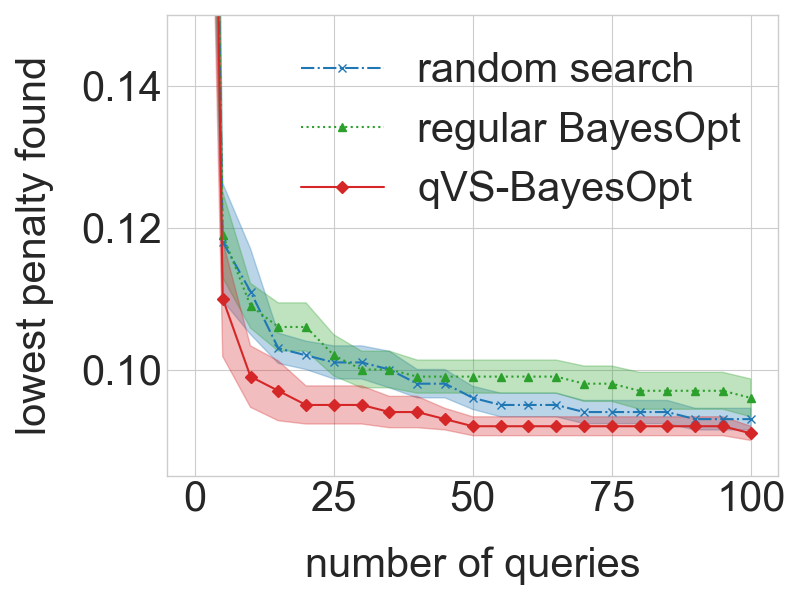}
\caption{
Energy penalty to release stored toxins (lower is better) of the best MOFs found by different algorithms. Our qVS-BayesOpt ($q = 1$) outperforms regular BayesOpt and random search. 
}
\label{fig:mof_efficiency}
}
\end{figure}

We further seek to visualize the search behaviors of our method by first using the multidimensional scaling technique \citep{cox2001multidimensional} to compute a 2-dimensional embedding from the kernel matrix of the candidates within the database.
This embedding is shown in the first panel of Fig. \ref{fig:mof_scatter}, where the scatter points' colors and opacity levels are set based on the corresponding MOFs' storage capacity levels (the objective value to be maximized).
We then mark the MOFs that are selected by each method in the remaining panels and observe a number of distinct trends: compared to random search, regular BayesOpt focuses on the lower-left region where the storage capacity is high; qVS-BayesOpt, on the other hand, further inspects the lower-right portion, which also contains high-capacity MOFs but in fewer numbers.
It is exactly this diverse sampling strategy that we hoped to achieve with the qVS.
We further compute a metric similar to the effective discovery count (EDC) in Section \ref{sec:experiments_as} whereby a MOF is classified as ``good'' if it yields a storage capacity (the objective value to be maximized) of at least $15$, and the VS of these good MOFs collected by each policy is reported.
We see that our qVS method outperforms both baselines; interestingly, regular BayesOpt yields a lower EDC than even random search---another indication of the failure mode of its overly exploitative strategy.
To study the effect of the order $q$ on the algorithm's performance, we include in Table \ref{tab:mof_varying_q} the best objective value found under $q \in  \{ 0, 0.1, 0.5, 1, 2, \infty \}$.
We see that $q = 0.1$ yields the best performance, while some values of $q$ lead to even worse performance than regular BayesOpt (when $q = 0$).
This behavior indicates the importance of setting $q$ appropriately; an interesting future direction could be to dynamically set $q$ based on search progress.

To conclude our analysis, we include another relevant metric in the MOF search application, which is the percentage energy penalty incurred when the stored toxins are eventually released for disposal, which ranges from 0 to 1 and a lower value indicates higher energy efficiency.
Here, Fig. \ref{fig:mof_efficiency} shows the average penalty across the optimization runs discussed above by the three algorithms, where we once again observe that (1) regular BayesOpt could fail to compete against even random search and (2) qVS-BayesOpt outperforms the two baselines.

 \glsresetall

\section{Conclusion}

We extended the Vendi scores to account for quality. We used these new quality-weighted Vendi scores, or qVS, to propose a unified framework for experimental design tasks to make diverse discoveries in discrete and continuous spaces. To optimize qVS, we proposed the sequential greedy strategy, widely used to optimize functions with diminishing returns. Our extensive experiments on scientific discovery problems show that the algorithms resulting from our framework can collect diverse, high-quality data, effectively balancing exploitation and exploration.

\subsection*{Acknowledgements}
Adji Bousso Dieng is supported by the National Science Foundation, Office of Advanced Cyberinfrastructure (OAC): \#2118201 and by Schmidt Sciences via the AI2050 Early Career Fellowship. 

\subsection*{Dedication}
This paper is dedicated to \href{https://en.m.wikipedia.org/wiki/Kwame_Nkrumah}{Kwame Nkrumah}.

\bibliographystyle{apa}
\bibliography{main}

\end{document}